\documentclass[10pt,twocolumn,letterpaper]{article}

\usepackage{wacv}
\usepackage{times}
\usepackage{epsfig}
\usepackage{graphicx}
\usepackage{amsmath}
\usepackage{amssymb}
\usepackage{multirow}

\DeclareMathOperator*{\argmax}{argmax}
\usepackage{mathtools}
\DeclarePairedDelimiterX{\infdivx}[2]{(}{)}{%
  #1\;\delimsize\|\;#2%
}
\newcommand{\infdiv}{D_{KL}\infdivx}

\makeatletter
\@namedef{ver@everyshi.sty}{}
\makeatother
\usepackage{tikz}
\usetikzlibrary{arrows}
\usepackage{verbatim}

\def\wacvPaperID{136} 

\wacvfinalcopy 

\ifwacvfinal
\fi

\ifwacvfinal
\usepackage[breaklinks=true,bookmarks=false]{hyperref}
\else
\usepackage[pagebackref=true,breaklinks=true,colorlinks,bookmarks=false]{hyperref}
\fi

\begin{document}

\title{Improving Model Generalization by Agreement of Learned Representations from Data Augmentation}

\author{Rowel Atienza\\
University of the Philippines\\
Electrical and Electronics Engineering Institute, Diliman, 1101 Quezon City, Philippines\\
{\tt\small rowel@eee.upd.edu.ph}

}

\maketitle

\begin{abstract}
Data augmentation reduces the generalization error by forcing a model to learn invariant representations given different transformations of the input image. In computer vision, on top of the standard image processing functions, data augmentation techniques based on regional dropout such as CutOut, MixUp, and CutMix and policy-based selection such as AutoAugment demonstrated state-of-the-art (SOTA) results. With an increasing number of data augmentation algorithms being proposed, the focus is always on optimizing the input-output mapping while not realizing that there might be an untapped value in the transformed images with the same label. We hypothesize that by forcing the representations of two transformations to agree, we can further reduce the model generalization error. We call our proposed method \textit{Agreement Maximization} or simply \textit{AgMax}. With this simple constraint applied during training, empirical results show that data augmentation algorithms can further improve the classification accuracy of ResNet50 on ImageNet by up to 1.5\%,  WideResNet40-2 on CIFAR10 by up to 0.7\%, WideResNet40-2 on CIFAR100 by up to 1.6\%, and LeNet5 on Speech Commands Dataset by up to 1.4\%. Experimental results further show that unlike other regularization terms such as label smoothing, AgMax can take advantage of the data augmentation to consistently improve model generalization by a significant margin. On downstream tasks such as object detection and segmentation on PascalVOC and COCO, \textit{AgMax} pre-trained models outperforms other data augmentation methods by as much as 1.0mAP (\textit{box}) and 0.5mAP (\textit{mask}). Code is available at \href{https://github.com/roatienza/agmax}{https://github.com/roatienza/agmax}. 
\end{abstract}

\tikzstyle{encoder}=[draw, fill=blue!20, minimum size=2em]
\tikzstyle{box}=[draw, minimum size=2em]
\tikzstyle{init} = [pin edge={to-,thin,black}]
\tikzstyle{image}=[minimum size=1.5em]

\begin{figure}[t]
\begin{center}

\begin{tikzpicture}[node distance=2cm,auto,>=latex']
    \node [encoder] (encoder) {$\mathcal{E}(\boldsymbol{x};\boldsymbol{\theta})$};
    
    \node (input) [left of=encoder,node distance=1.5cm, coordinate] {encoder};
    \node [image] (image) [left of=input,node distance=0.8cm] {\includegraphics[width=.1\textwidth]{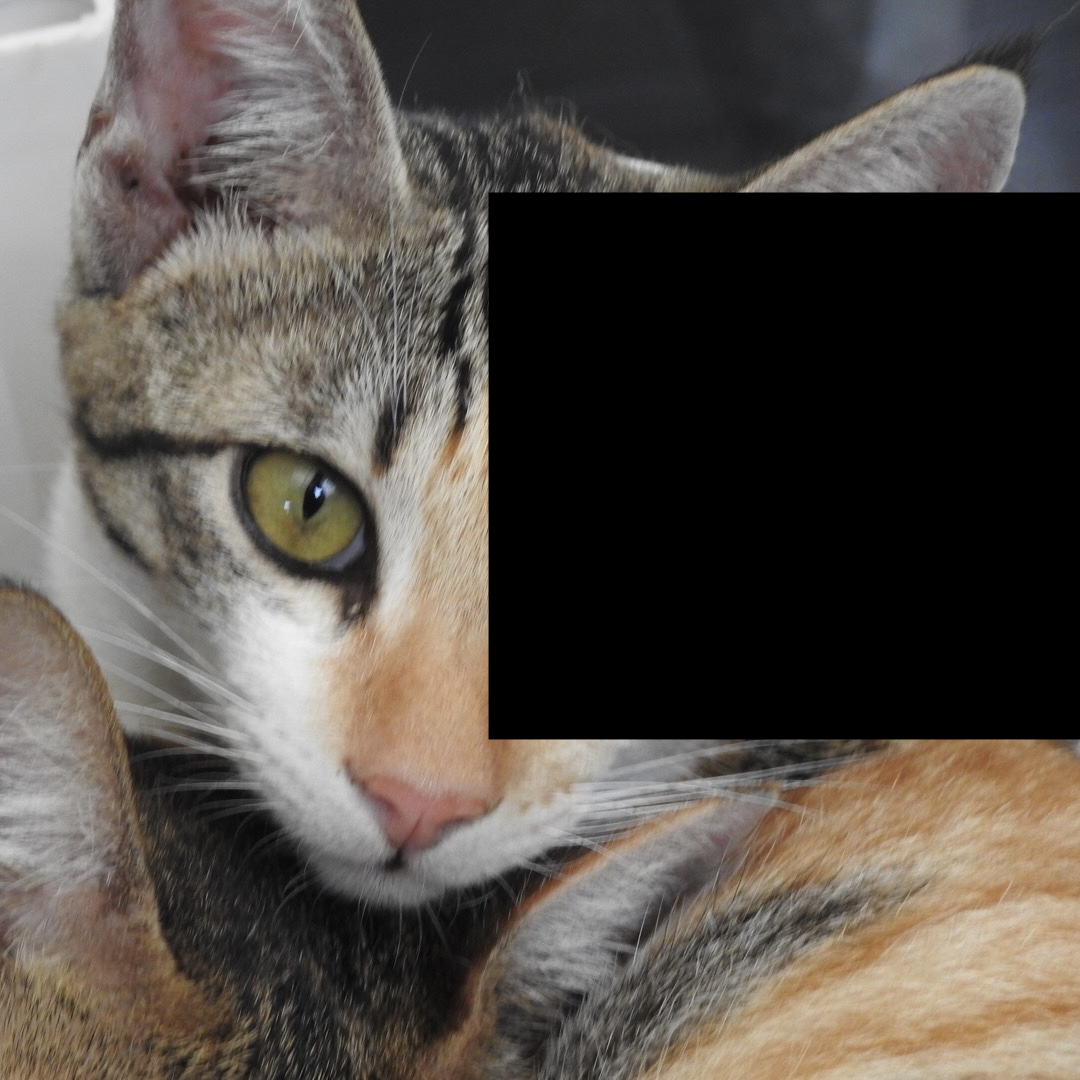}};
    \node [box] (softmax) [right of=encoder] {$\sigma(\boldsymbol{z})$};
    \node [coordinate] (output) [right of=softmax, node distance=1.5cm]{};
    \path[->] (input) edge node {$\boldsymbol{x}$} (encoder);
    \path[->] (encoder) edge node {$\boldsymbol{z}$} (softmax);
    \draw[->] (softmax) edge node {$p(\hat{y})$} (output) ;
    \node (output_label) [right of=softmax,node distance=0.9cm, coordinate] {};
    \node [below of=output_label,node distance=.2cm] {"cat"};
\end{tikzpicture}

\end{center}
   \caption{In supervised learning, a model $\mathcal{E}$ is trained to find the optimal parameters $\boldsymbol{\theta^*}$. Input data augmentation improves the performance by forcing $\mathcal{E}$ to learn invariant representations $\boldsymbol{z}$ under image transformation such as removing a random square region as done in CutOut. $\boldsymbol{x}$ is a labelled positive sample.}
\label{fig:single_network}
\end{figure}
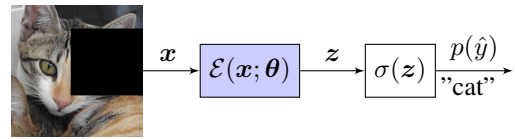

\begin{figure}
\begin{center}

\begin{tikzpicture}[node distance=2.5cm,auto,>=latex']
    \node [encoder] (encoder1) {$\mathcal{E}(\boldsymbol{x}_{1};\boldsymbol{\theta})$};
    \node (input1) [left of=encoder1,node distance=1.5cm, coordinate] {encoder1};
    \node [image] (image1) [left of=input1,node distance=0.8cm] {\includegraphics[width=.1\textwidth]{images/cat_cutout_1.png}};
    \node [box] (softmax1) [right of=encoder1] {$\sigma(\boldsymbol{z}_1)$};
    \node [coordinate] (output1) [right of=softmax1, node distance=1.7cm]{};
    \path[->] (input1) edge node [below] {$\boldsymbol{x}_{1}$} (encoder1);
    \path[->] (encoder1) edge node [name=z1, below]{$\boldsymbol{z}_{1}$} (softmax1);
    \path[->] (softmax1) edge node  [above]{$p(\hat{y}_{1})$} (output1) ;
    \node (output_label1) [right of=softmax1,node distance=1cm, coordinate] {};
    \node [below of=output_label1,node distance=.2cm] {"cat"};

    \node [encoder] (encoder2) [above of=encoder1] {$\mathcal{E}(\boldsymbol{x}_{2};\boldsymbol{\theta})$};
    \node (input2) [left of=encoder2,node distance=1.5cm, coordinate] {encoder2};
    \node [image] (image2) [left of=input2,node distance=0.8cm] {\includegraphics[width=.1\textwidth]{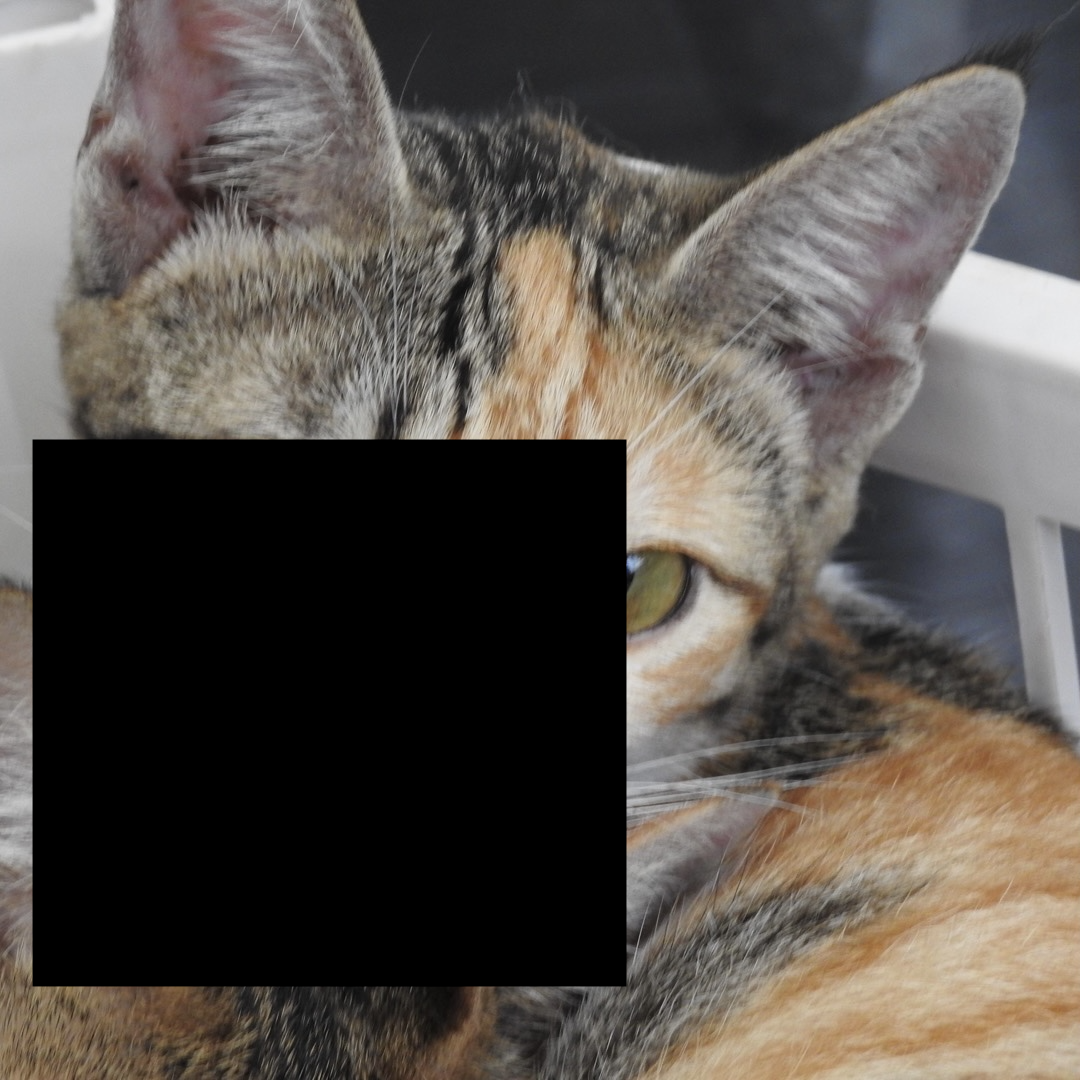}};
    \node [box] (softmax2) [right of=encoder2] {$\sigma(\boldsymbol{z}_2)$};
    \node [coordinate] (output2) [right of=softmax2, node distance=1.7cm]{};

    \node (agreement) [above of=z1, node distance=1.5cm] {Agreement};  
    
    \node (positive) [above of=image1, node distance=1.2cm] {2 positive samples};  

    \path[->] (input2) edge node {$\boldsymbol{x}_{2}$} (encoder2);
    \path[->] (encoder2) edge node [name=z2]{$\boldsymbol{z}_{2}$} (softmax2);
    \path[<->] (encoder1) edge node {$\boldsymbol{\theta}$} (encoder2);
    \path[->] (z1) edge node {} (agreement); 
    \path[->] (z2) edge node {} (agreement);
    \path[->] (softmax2) edge node {$p(\hat{y}_{2})$} (output2) ;
    \node (output_label2) [right of=softmax2,node distance=1cm, coordinate] {};
    \node [below of=output_label2,node distance=.2cm] {"cat"};    
    
    \node [below of=output_label2,node distance=1.2cm] {Same label};

\end{tikzpicture}

\end{center}
   \caption{In supervised learning with \textit{AgMax}, we impose an additional constraint that representations $\boldsymbol{z}_{1}$ and $\boldsymbol{z}_{2}$ must also agree. The two parallel models are just one and the same and share the same set of parameters $\boldsymbol{\theta}$. $\boldsymbol{x}_1$ and $\boldsymbol{x}_2$ are 2 positive samples with the same label.}
\label{fig:agmax_network}
\end{figure}

\section{Introduction}

\begin{table}[]
    \centering
   
    \begin{tabular}{l |c | c | c}
     \hline
     Data & {} & with Label &  with  \\
    Augmentation & Baseline & Smoothing & AgMax  \\
     \hline
    Standard & $76.4\pm0.1$ & $76.8\pm0.1$ & $76.9\pm0.1$ \\
    CutOut\cite{devries2017improved} & $76.2\pm{0.0}$ & $76.5\pm{0.1}$  & $77.1\pm{0.0}$ \\
    MixUp\cite{zhang2017mixup} & $76.5\pm{0.1}$ & $76.7\pm{0.1}$ & $\boldsymbol{77.6}\pm{0.1}$\\
    CutMix\cite{yun2019cutmix} & $76.3\pm{0.0}$ & $76.4\pm{0.1}$ & $77.4\pm{0.0}$\\
    AA\cite{cubuk2019autoaugment} & $76.2\pm{0.1}$ & $76.2\pm{0.1}$  & $77.1\pm{0.1}$\\
    CutOut+AA\cite{cubuk2019autoaugment} & $75.7\pm{0.1}$ & $75.7\pm{0.1}$ & $76.6\pm{0.1}$\\

    MixUp+AA & $75.9\pm{0.0}$ & $76.5\pm{0.1}$ & $77.1\pm{0.1}$ \\
    CutMix+AA & $75.5\pm{0.1}$ & $75.5\pm{0.1}$ & $77.0\pm{0.1}$\\
     \hline
    \end{tabular}
    \caption{Top-1\% accuracy of ResNet50 trained for 90 epochs on ImageNet using different data augmentation methods with Label Smoothing or AgMax. Standard data augmentation is random horizontal flipping, color jitter  and lighting. }
    \label{tab:smooth_vs_agmax}
\end{table}

\begin{table*}
\small
\begin{center}
\begin{tabular}{l | c |  r r  r r  r r r r  }

\hline
{} & {} &  \multicolumn{2}{c }{CIFAR10\cite{krizhevsky2009learning}} & \multicolumn{2}{c }{CIFAR100 } & \multicolumn{4}{c }{ImageNet \cite{russakovsky2015imagenet}}\\

{} & {} &  \multicolumn{4}{c }{WideResNet \cite{zagoruyko2016wide}}  & \multicolumn{4}{c }{ResNet50 \cite{he2016deep}}\\

Data & {} & \multicolumn{4}{c }{200 epochs}  & \multicolumn{2}{c }{90 epochs} & \multicolumn{2}{c }{270 epochs} \\

Augmentation & Year & 40-2 & 28-10 & 40-2 & 28-10 & Top-1 & Top-5 & Top-1 & Top-5  \\
\hline
\multicolumn{10}{c }{\textbf{Published Results}}\\
\hline
Standard & - &  94.7 & 96.1 & 74.0 & 81.2 & 76.2 & 92.9 & 76.3 & 93.1  \\
CutOut\cite{devries2017improved} & 2017  & 95.9 & 96.9 & 74.8 & 81.6 & - & - & 77.1 & 93.3   \\
MixUp\cite{zhang2017mixup} & 2018  &  - & 97.3 & - & 82.5 & 76.7 & 93.4 & 77.9 & 93.9    \\
CutMix\cite{yun2019cutmix} & 2019  &  - & - & - & - & - & - & 78.6 & 94.1    \\

AA\cite{cubuk2019autoaugment} & 2019  &  - & - & - & - & - & - & 77.6 & 93.8   \\
CutOut+AA\cite{cubuk2019autoaugment} & 2019 &  96.3 & 97.4 & 79.3 & 82.9 & - & - & - & -    \\
\hline
\multicolumn{10}{c }{\textbf{without AgMax} (Reproduced Results)} \\
\hline
Standard & - & 95.1 & 96.2 & 76.9 & 81.3 & 76.4 & 93.2 & 76.8 & 93.3    \\
CutOut\cite{devries2017improved} & 2017  & 96.2 & 97.1 & 78.4 & 82.5 & 76.2 & 93.1 & 77.5 & 93.6   \\
MixUp\cite{zhang2017mixup} & 2018  & 95.8 & 97.1 & 78.3 & 82.8 & \underline{76.5} & \underline{93.3} & 78.2 & 93.9    \\
CutMix\cite{yun2019cutmix} & 2019 &  96.2 & 97.3 & 79.4 & 83.8 & 76.3 & 93.2 & \underline{78.7} & \underline{94.2}    \\
AA\cite{cubuk2019autoaugment} & 2019 & 95.9 & 96.9 & 78.4 & 82.8 & 76.2 & 93.1 & 77.6 & 93.6   \\
CutOut+AA\cite{cubuk2019autoaugment} & 2019  &  \underline{96.4} & \underline{97.5} & 80.0 & 83.8 & 75.7 & 92.8 & 77.9 & 93.7   \\
MixUp+AA & - &   96.0 & 97.4 & 79.1 & 84.1 & 75.9 & 92.9 & 78.3 & 94.0   \\
CutMix+AA & - &  \underline{96.4} & 97.4 & \underline{80.1} & \underline{85.0} & 75.5 & 92.7 & 78.5 & 94.1   \\
\hline
\multicolumn{10}{c }{\textbf{with AgMax}}\\
\hline
Standard & - &  95.6(0.5) & 96.4(0.2) & 77.4(0.5) & 81.7(0.4) & 76.9(0.5) & 93.5(0.3) & 77.2(0.4) & 93.6\textbf{(0.3)}    \\
CutOut & - &  96.6(0.4) & 97.3(0.2) & 79.2(0.8) & 82.6(0.1) & 77.1(0.9) & 93.6(0.5) & 77.6(0.1) & 93.8(0.2)    \\
MixUp & - &  96.3(0.5) & 97.5(0.4) & 79.5(1.2) & 82.9(0.1) & \textbf{77.6}(1.1) & 93.7(0.4) & 78.4(0.2) & 94.1(0.2)    \\
CutMix & - & 96.7(0.5) & 97.7(0.4) & 80.0(0.6) & 84.0(0.2) & 77.4(1.1) & \textbf{93.9}(0.7) & 79.0(0.3) & 94.2(0.0)   \\
AA & - &  96.4(0.5) & 97.4(\textbf{0.5}) & 79.2(0.8) & 83.0(0.2) & 77.1(0.9) & 93.4(0.3) & 77.7(0.1) & 93.8(0.2)     \\

CutOut+AA & - &  \textbf{97.1(0.7)} & 97.8(0.3) & 81.1(1.1) & 84.0(0.2) & 76.6(0.9) & 93.4(0.6) & 78.2(0.3) & 94.0\textbf{(0.3)}    \\

MixUp+AA & - &  96.6(0.6) &\textbf{97.9(0.5)}  & 80.7(\textbf{1.6}) & 84.8\textbf{(0.7)} & 77.1(1.2) & 93.5(0.6) & 78.6(0.3) & 94.2(0.2)    \\
CutMix+AA & - &  96.8(0.4) & 97.8(0.4) & \textbf{81.3}(1.2) & \textbf{85.3}(0.3) & 77.0\textbf{(1.5)} & 93.5\textbf{(0.8)} & \textbf{79.1(0.6)} & \textbf{94.4(0.3)}   \\

\hline
\end{tabular}
\vskip 0.1in
\caption{Evaluation landscape showing model accuracy of different data augmentation algorithms with and without \textit{AgMax}. Underscore is the best performing configuration without \textit{AgMax}. Bold is the best performing method for all configurations. The absolute percentage increase in accuracy due to \textit{AgMax} is enclosed in parentheses. Standard data augmentation algorithm is defined in the Experimental Results section.}
\label{tab:data_aug_eval}
\end{center}
\vskip -0.1in
\end{table*}

\begin{table}
\small
\begin{center}
\begin{tabular}{l |  r r r r}

\hline
{} &  \multicolumn{4}{c }{Speech Commands \cite{warden2018speech}}\\

{} &   \multicolumn{2}{c }{LeNet5 \cite{lecun1998gradient}} & \multicolumn{2}{c }{VGG11 \cite{simonyan2014very}} \\

{Data} &   \multicolumn{4}{c }{30 epochs}\\

Augmentation  & test & val & test & val\\
\hline
\multicolumn{5}{c }{\textbf{Published Results}}\\
\hline
Standard & 89.7 & 90.2 & 95.4 & 95.0  \\
CutOut\cite{devries2017improved}  & - & - & - & -  \\
MixUp\cite{zhang2017mixup}  &  89.2 & 89.9 & 96.6 & 96.1  \\
CutMix\cite{yun2019cutmix}  &  - & - & - & -  \\
\hline
\multicolumn{5}{c }{\textbf{without AgMax} (Reproduced Results)}\\
\hline
Standard  & \underline{89.9} & \underline{90.0} & 96.3 & 96.0  \\
CutOut\cite{devries2017improved}  & 89.0 & 89.2 & \underline{96.5} & 96.1  \\
MixUp\cite{zhang2017mixup} & 89.4 & 89.4 & \underline{96.5} & \underline{96.2}  \\
CutMix\cite{yun2019cutmix}  & 87.1 & 87.5 & 96.4 & \underline{96.2}  \\
\hline
\multicolumn{5}{c }{\textbf{with AgMax}}\\
\hline
Standard  & 90.2(0.3) & \textbf{90.0}(0.0) & 96.4(0.1) & 96.1(0.1)  \\
CutOut &  \textbf{90.4}(1.4) & \textbf{90.0}(0.8) & 96.5(0.0) & 96.1(0.0)  \\
MixUp  & 89.4(0.0) & 89.6(0.2) & \textbf{96.8(0.3)} & 96.3(0.1)  \\
CutMix  & 88.8\textbf{(1.7)} & 89.3\textbf{(1.8)} & 96.7\textbf{(0.3)} & \textbf{96.4(0.2)}  \\

\hline
\end{tabular}
\vskip 0.1in
\caption{Evaluation landscape showing Top-1\% model accuracy of different data augmentation configurations on Speech Commands Dataset. AutoAugment is not included since there is no publicly available policy for Speech Commands Dataset.}
\label{tab:speech_data_aug_eval}
\end{center}
\vskip -0.1in
\end{table}

To achieve state-of-the-art (SOTA) performance, data augmentation plays a crucial role in both supervised and self-supervised model training. In computer vision, image processing functions such as rotation, translation, cropping, flipping, and color distortion improve model generalization. In recent years, a strong interest in new data augmentation techniques has emerged because of the significant improvement in model performance compared to baseline scores.

Instead of just applying random image processing operations, policy-based methods such as AutoAugment (AA) \cite{cubuk2019autoaugment}, FastAugment \cite{lim2019fast}, RandAugment \cite{cubuk2020randaugmentneurips}, Adversarial AutoAugment \cite{zhang2019adversarial}, and PBA \cite{ho2019population} carefully select a recipe of image operations to generate new input data to minimize model overfitting. Regional dropouts or techniques based on direct image alteration such as CutOut \cite{devries2017improved}, RICAP \cite{takahashi2018ricap}, CutMix \cite{yun2019cutmix}, GridMask \cite{chen2020gridmask} and MixUp \cite{zhang2017mixup} improve model performance by forcing the model to learn invariant representations.

With an increasing number of data augmentation algorithms being introduced, as shown in Figure \ref{fig:single_network} the focus is always on optimizing the input-output mapping. We hypothesize that there might be  an untapped value between representations of transformed inputs with the same label. Basic intuition tells us that under different transformations such as in Figure \ref{fig:agmax_network}, two inputs with the same label should agree on which representations that a model learns. These two inputs are called \textbf{positive samples} since they have the same label but two different transformations. A classifier receiving two positive samples of a \textit{cat} must learn to extract the \textit{minimum common set of representations} such as the presence of whiskers, fur, sharp eyes, short nose, etc. We call this function \textit{Agreement}. In this paper, we use mutual information (MI) to estimate \textit{Agreement}. We performed ablation studies to demonstrate that other agreement functions such as MSE, KL-divergence and cross-entropy (CE) are also effective.

Using a common evaluation protocol, experimental results indicate that our proposed method \textit{AgMax} improves the performance of almost all model-dataset configurations. On ResNet50 trained on ImageNet for 90 epochs, as shown in Table \ref{tab:smooth_vs_agmax}, \textit{AgMax}  consistently outperforms Label Smoothing \cite{szegedy2016rethinking} especially under heavy data augmentation. In a bigger evaluation landscape, the results in Tables \ref{tab:data_aug_eval} and \ref{tab:speech_data_aug_eval} demonstrate the consistent improvement in generalization for different models and datasets due to \textit{AgMax}. On downstream tasks such as object detection and segmentation, a ResNet50 model pre-trained with \textit{AgMax} outperforms its counterpart pre-trained model by as much as 1.0mAP on \textit{bounding box} and 0.5mAP on \textit{segmentation mask}.

\section{Related Work}
To achieve state-of-the-art (SOTA) performance, data augmentation plays a crucial role in both supervised and self-supervised model training. Data augmentation belongs to a bigger field of study called regularization. The objective of regularization is to improve model generalization by modifying the network structure during training, augmenting the train dataset, modifying the loss function or modifying the model training algorithm. For example, dropout \cite{srivastava2014dropout} randomly drops neural network units during training to mimic data and network perturbations. As a result, a model improves its test performance. In deep CNNs, instead of dropping feature maps, noise injection or substitution such as Stochastic Depth \cite{huang2016deep}, Shake-Shake \cite{gastaldi2017shake}, DropBlock \cite{ghiasi2018dropblock}, DropPath \cite{larsson2016fractalnet}, SpatialDropout \cite{tompson2015efficient} and Shake-Drop \cite{yamada2019shakedrop} are used.  Related to dropout is the regional dropout. Instead of dropping neural network units, a certain region of the input is removed, mixed or blended. In effect, regional dropout augments the training dataset by exposing a model to extreme input data transformations. CutOut \cite{devries2017improved}, RICAP \cite{takahashi2018ricap}, CutMix \cite{yun2019cutmix}, GridMask \cite{chen2020gridmask} and MixUp \cite{zhang2017mixup} belong to this category. 



Before the regional dropout methods were proposed, data augmentation was achieved by basic input data transformations. In computer vision,  padding, random cropping, translation, rotation, horizontal flipping and color distortion are commonly used. Recently, these standard data augmentation techniques have been supplanted by a more structured learned policy in order to arrive at an optimal recipe of data transformation functions. In computer vision, AutoAugment \cite{cubuk2019autoaugment}, FastAugment \cite{lim2019fast}, Adversarial AutoAugment \cite{zhang2019adversarial}, RandAugment \cite{cubuk2020randaugmentneurips} and PBA \cite{ho2019population} have been proposed.  Among the available data augmentation methods, regional dropouts, policy-based, and gradient augmentation have demonstrated state-of-the-art results. Improving model generalization can also be achieved by modifying the loss function. Label smoothing \cite{szegedy2016rethinking} and weight decay \cite{hanson1989comparing} belong to this category. 

In this paper, the idea is to take advantage of the huge amount of data generated by augmentation methods. If an image undergoes data augmentation to produce 2 new images, the resulting representations a model learns must agree since both inputs belong to the same category. This idea bears resemblance to representations matching in self-supervised learning such as BYOL \cite{NEURIPS2020_f3ada80d} and DINO \cite{caron2021emerging} where the predictions of teacher(target) and student(online) networks on 2 positive samples (2 views of the same image) are reinforced by similarity learning. The similarity function could be MSE in BYOL or cross-entropy (CE) function in DINO. The teacher and student networks could be similar in architecture but have different sets of parameters. The teacher network parameters are exponential moving average of the online parameters. The main difference of \textit{AgMax} is it uses one network for both predictions. Instead of maximizing a similarity function, \textit{AgMax} maximizes the agreement using mutual information.     

In this paper, we validate our hypothesis on 4 commonly used data augmentation algorithms: 1 policy-based approach and 3 regional dropout techniques. We chose AutoAugment for the policy-based data augmentation given that its policy is publicly available. FastAugment, Adversarial AutoAugment, and RandAugment are built on top of the key ideas of AutoAugment. For regional dropout, we used CutOut, MixUp, and CutMix. These methods have gained mainstream use and achieved state-of-the-art results. Furthermore, their code implementations are publicly available for reproducibility. 

\section{Improving Generalization by Agreement}

With reference to Figure \ref{fig:agmax_network}, we hypothesize that the representations of two images derived by applying a data augmentation method on an image must agree for a model to further improve its classification accuracy. In this paper, we propose that the \textit{Agreement} is the amount of shared information between the two views of the same image. Therefore, on top of the classification loss function,  maximizing the mutual information between the two representations could improve the model generalization. The total loss function can be expressed as:

\begin{align}\label{eq:new_loss_function}
\mathcal{L} = \mathcal{L}_{CE} + \lambda\mathcal{L}_{MI}.
\end{align}
    
$\lambda$ is the weight of the MI loss function. 

 For the case of discrete random variables such as in image classification, mutual information is expressed as:

\begin{equation}\label{eq:mi_eq}
      I(\boldsymbol{z}_{1};\boldsymbol{z}_{2})=\sum_{\boldsymbol{z}_{1},\boldsymbol{z}_{2}}{P(\boldsymbol{z}_{1},\boldsymbol{z}_{2})}\log\frac{P(\boldsymbol{z}_{1},\boldsymbol{z}_{2})}{P(\boldsymbol{z}_{1})P(\boldsymbol{z}_{2})}.
\end{equation}

In other words, MI is the KL-divergence between the joint and product of marginal probabilities of $\boldsymbol{z}_{1}$ and $\boldsymbol{z}_{2}$.

In recent years, several neural MI estimators have been proposed \cite{belghazi2018mine,tschannen2019mutual,hjelm2018learning,ji2019invariant,nguyen2010estimating}. Invariant Information Clustering (IIC) \cite{ji2019invariant} proposed a method to estimate Equation \ref{eq:mi_eq}.

For a given dataset or batch of size $n$, the joint probability matrix $\boldsymbol{P}\in\mathbb{R}^{C\times{C}}$ can be computed as:

\begin{equation}
    \boldsymbol{P} = \frac{1}{n}\sum_{i=1}^n\Phi(\boldsymbol{x}_{1,i})\cdot{\Phi(\boldsymbol{x}_{2,i})}^\top,
\end{equation}

where $\boldsymbol{x}_{1,i}$ and $\boldsymbol{x}_{2,i}$ are two transformed versions of the same image $\boldsymbol{x}_{i}$. $\Phi(\boldsymbol{x})=\sigma(\mathcal{E}(\boldsymbol{x}))=softmax(\boldsymbol{z})\in{[0,1]^C}$. This can be interpreted as the distribution of $z$ over $C$ classes formally given as $P(z=c |{\boldsymbol{x},\boldsymbol{\theta}}) )=\Phi_c(\boldsymbol{x})$. The marginal distributions $\boldsymbol{P}_{c_1}=P(z_1=c_1)$ and $\boldsymbol{P}_{c_2}=P(z_2=c_2)$ can be obtained by summing the rows and columns of $\boldsymbol{P}$ respectively. Each element of $\boldsymbol{P}$ is the joint probability $\boldsymbol{P}_{{c_1}{c_2}}=P(z_1=c_1,z_2=c_2)$. Since $\boldsymbol{P}_{{c_1}{c_2}}=\boldsymbol{P}_{{c_2}{c_1}}$, the matrix $\boldsymbol{P}$ must be symmetric. Ensuring a symmetric $\boldsymbol{P}$ is done by $\boldsymbol{P}=\frac{\boldsymbol{P}+\boldsymbol{P}^\top}{2}$.    

Using the joint and marginal probabilities, the mutual information loss is computed as:

\begin{align}\label{eq:mi_loss_function}
\mathcal{L}_{MI} = -I(\boldsymbol{z}_1;\boldsymbol{z}_2) = -\sum_{c_1=1}^C\sum_{c_2=1}^C{\boldsymbol{P}_{{c_1}{c_2}}}\ln\frac{\boldsymbol{P}_{{c_1}{c_2}}}{\boldsymbol{P}_{{c_1}}\cdot\boldsymbol{P}_{{c_2}}}.
\end{align}

A 2-layer MLP network can also be used to estimate the joint distribution. The MLP network is trained using the objective functions:  $p(\boldsymbol{z}_{1},\boldsymbol{z}_{2}) \xrightarrow{}p(\boldsymbol{z}_{1},\boldsymbol{z}_{1})$ and  $p(\boldsymbol{z}_{1},\boldsymbol{z}_{2})\xrightarrow{}p(\boldsymbol{z}_{2},\boldsymbol{z}_{2})$ since both features refer to the same class and so is the joint distribution.

Note that to estimate the MI loss, only pairs of positive samples or one-to-one mapping is needed. Therefore, \textit{AgMax} works even for small batch sizes. This is different from constrastive learning that requires a positive and many negative samples or one-to-many mapping. Contrastive learning needs large batch sizes (\eg 4,096 and up) to work. This has a huge negative implication on GPU memory requirements.  


\subsection{Agreement by Mutual Information}

In this section, we \textbf{attempt} to find a possible explanation on why MI provides a good \textit{Agreement} function using the Maximum a Posteriori (MAP) principle:

\begin{align}\label{eq:map}
    \boldsymbol{{\theta}^{*}} =
    \argmax_{\boldsymbol{\theta}}\log{p(\boldsymbol{\theta}|\mathcal{D})}=     \argmax_{\boldsymbol{\theta}}\log{p(\boldsymbol{\theta}|{\boldsymbol{x},\boldsymbol{y}})}.
\end{align}

When applied to deep neural networks (DNNs) as shown in Figure \ref{fig:single_network}, $\boldsymbol{\theta}$ is the model parameters while $\boldsymbol{\theta^{*}}$ represents the maximal point estimate for a given dataset $\mathcal{D}=\{\boldsymbol{x},\boldsymbol{y}\}$. 

Using Bayes' Theorem and with the constant term dropped, the conditional probability in Equation \ref{eq:map} can be rewritten as:

\begin{align}\label{eq:p_xtheta}
     \boldsymbol{\theta^{*}}=\argmax_{\boldsymbol{\theta}}\left[\log{p(\boldsymbol{y}|{\boldsymbol{x},\boldsymbol{\theta}})}+\log{p(\boldsymbol{x}|{\boldsymbol{\theta}})p({\boldsymbol{\theta}})}\right].
\end{align}

In a given model, a backbone network encodes the input into a latent variable, $\boldsymbol{z}=\mathcal{E}(\boldsymbol{x};\boldsymbol{\theta}_{1})$. A decoder decides what output is generated, $\hat{\boldsymbol{y}}=\mathcal{D}(\boldsymbol{z};\boldsymbol{\theta}_{2})$. Collectively, $\boldsymbol{\theta}=\{\boldsymbol{\theta}_{1}, \boldsymbol{\theta}_{2}\}$. If $\boldsymbol{z}$ is taken from the last layer before the non-parametric softmax prediction of $\boldsymbol{\hat{y}}=\sigma(\boldsymbol{z})$, then $\boldsymbol{\theta}=\{\boldsymbol{\theta}_{1},\varnothing\}$ and $\boldsymbol{\hat{y}}$ is a good proxy of $\boldsymbol{z}$.  During supervised training, the goal is to estimate the empirical distribution $p(\boldsymbol{y}|\boldsymbol{x})$ using a parameterized distribution $p(\boldsymbol{\hat{y}}|{\boldsymbol{x},\boldsymbol{\theta}})$ as modelled by the encoder-decoder. This is done by minimizing a distance function such as the Kullback-Leibler divergence function $\infdiv{p(\boldsymbol{y}|\boldsymbol{x})}{p(\boldsymbol{\hat{y}}|{\boldsymbol{x},\boldsymbol{\theta}})}$. In supervised classification problems, this is equivalent to minimizing the categorical cross-entropy loss function, $\mathcal{L}_{CE}=-\mathbb{E}_{p(\boldsymbol{y}|\boldsymbol{x})}\log{p(\boldsymbol{\hat{y}}|{\boldsymbol{x},\boldsymbol{\theta}})}$.

As shown in Figure \ref{fig:agmax_network}, given two positive samples of $\boldsymbol{x}$, we can reformulate MAP as:
\begin{multline}\label{eq:theta_expectation}
       \boldsymbol{\theta^{*}} = \argmax_{\boldsymbol{\theta}} 
       \left[ 
       \frac{1}{2}\sum_{i=1}^2 \mathbb{E}\log{p(\boldsymbol{y}|{\boldsymbol{x}_{i},\boldsymbol{\theta}})} + \right. \\
       \left. \frac{1}{2}\mathbb{E}\Big[\log{\frac{ p(\boldsymbol{z}_{1},\boldsymbol{z}_{2}| \boldsymbol{x}_{1}, \boldsymbol{x}_{2}, \boldsymbol{\theta}) }{ p(\boldsymbol{z}_{1}| \boldsymbol{x}_{1}, \boldsymbol{\theta})p(\boldsymbol{z}_{2}| \boldsymbol{x}_{2}, \boldsymbol{\theta}) }}\Big]+\frac{1}{2}\mathbb{E}\log{p(\boldsymbol{x}_{1},\boldsymbol{x}_{2},\boldsymbol{\theta})} 
        \right],  
\end{multline}

since Equation \ref{eq:p_xtheta} can be rewritten as:

\begin{multline}\label{eq:p_xtheta_with_z}
       \boldsymbol{\theta^{*}}= \argmax_{\boldsymbol{\theta}} \left[  \log{p(\boldsymbol{y}|{\boldsymbol{x},\boldsymbol{\theta}})} + \log{ p(\boldsymbol{z}, \boldsymbol{x}, \boldsymbol{\theta}) } \right. \\
        \left. - \log{ p(\boldsymbol{z}| \boldsymbol{x}, \boldsymbol{\theta}) } \right].  
\end{multline}

Given two data augmentations for a given input: 
\begin{multline}\label{eq:p_xtheta_with_2z}
       \boldsymbol{\theta^{*}}=
       \argmax_{\boldsymbol{\theta}} \left[ 
       \frac{1}{2}\sum_{i=1}^2 \log{p(\boldsymbol{y}|{\boldsymbol{x}_{i},\boldsymbol{\theta}})} + \right. \\
       \left.
       \frac{1}{2}\log{ p(\boldsymbol{z}_{1}, \boldsymbol{x}_{1}, \boldsymbol{\theta})p(\boldsymbol{z}_{2}, \boldsymbol{x}_{2}, \boldsymbol{\theta}) }
        - \frac{1}{2}\sum_{i=1}^2 
       \log{ p(\boldsymbol{z}_{i}| \boldsymbol{x}_{i}, \boldsymbol{\theta}) } \right],  
\end{multline}
Assuming independence, the second term in Equation \ref{eq:p_xtheta_with_2z} can be expressed as:
\begin{multline}\label{eq:upperbound_of_pz}
       \log{ p(\boldsymbol{z}_{1}, \boldsymbol{x}_{1}, \boldsymbol{\theta})p(\boldsymbol{z}_{2}, \boldsymbol{x}_{2}, \boldsymbol{\theta}) } = \\
       \log{ p(\boldsymbol{z}_{1}, \boldsymbol{z}_{2}, \boldsymbol{x}_{1}, \boldsymbol{x}_{2}, \boldsymbol{\theta}) }.  
\end{multline}

With dataset sampling, Equation \ref{eq:p_xtheta_with_2z} can be rewritten as Equation \ref{eq:theta_expectation}. 
The second term in Equation \ref{eq:theta_expectation} resembles the mutual information (MI): $I(\boldsymbol{z}_{1};\boldsymbol{z}_{2}|\boldsymbol{x}_{1},\boldsymbol{x}_{2},\boldsymbol{\theta})$. Maximizing the MI between the representations of $\boldsymbol{x}_{1}$ and $\boldsymbol{x}_{2}$ regularizes the choice of model parameters $\boldsymbol{\theta}$. In Figure \ref{fig:agmax_network} for example, the network is encouraged to find the representations of two views of the same \textit{cat}, such that the shared information between $\boldsymbol{z}_{1}$ and $\boldsymbol{z}_{2}$ is maximized. This could be finding the common features such as presence of whiskers,  fur,  sharp eyes,  short nose,  etc.

The third term in Equation \ref{eq:theta_expectation} can be rewritten as $\frac{1}{2}\mathbb{E}\log{p(\boldsymbol{\theta})}$ since the model parameters can be assumed to be independent from the input distribution. This term can be represented by $L2$ weight regularization. The first term of Equation \ref{eq:theta_expectation} can be optimized by the standard cross entropy loss function as discussed earlier in this section. Therefore, the total loss function can be written as Equation \ref{eq:new_loss_function}.

    

\begin{figure}
\centering

        \includegraphics[scale=0.51]{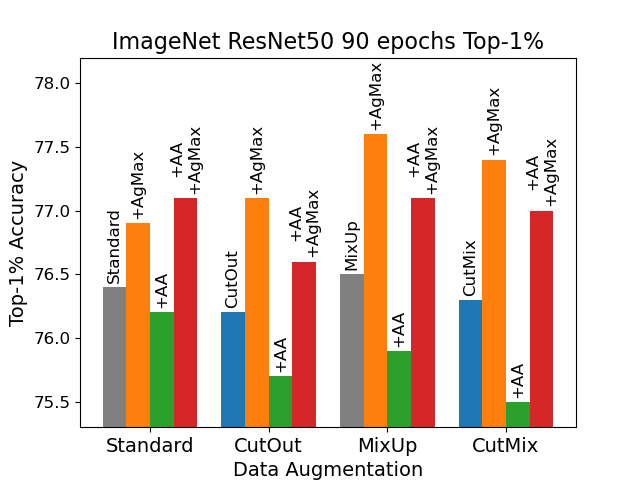}

 \caption{Top-1\% accuracy of different regularizer configurations on ImageNet using ResNet50 trained for 90 epochs.}
\label{fig:imagenet_resnet50_90_270_top1}
\end{figure}


\section{Experimental Results}

To validate our hypothesis, we first reproduced different data augmentation results. Then, we applied \textit{AgMax} to verify if the model generalization will improve.

To arrive at a fair comparison, we looked for the common evaluation protocols among the data augmentation algorithms under study starting with datasets and encoders or backbone networks. CIFAR10, CIFAR100 and ImageNet datasets are commonly used. We included Speech Commands Dataset as evaluated by MixUp. Except for CutOut, all methods used ResNet50 as the backbone network on ImageNet. We included WideResNet28-10 and  WideResNet40-2 as used by AutoAugment, RandAugment, and MixUp on CIFAR10, and CIFAR100. Lastly, we trained VGG11 and LeNet5 on Speech Commands Dataset as done in MixUp. 

However, an analysis of published experiments revealed that it is difficult to make a fair comparison of scores produced by different data augmentation algorithms due to a lack of consistent evaluation protocol. For example, on ImageNet dataset, the ResNet50 model was trained for epochs $ep{=}90$ and  $200$, $lr{=}0.4$, $bs{=}1,024$ in MixUp, and  $ep{=}300$, $lr{=}0.1$, $bs{=}256$ in CutMix. It leads to an unfair comparison since the performance of ResNet varies with the number of epochs, batch size, and learning rate settings. To address these issues, we formulated a common training condition. Then, we reproduced the reported scores. In the following subsections, we discuss the details of the  uniform experimental setup that we used for each model and dataset. Whenever possible, we implemented the settings in the published literature or official code implementations.

A further examination of published experiments showed that policy-based methods are seldom evaluated with complementary regional dropout algorithms. For example, AutoAugment has demonstrated that it can achieve better results with CutOut but its use with other regional dropout algorithms was not fully exploited. In the Published Results section of Table \ref{tab:data_aug_eval}, only 58\% of the evaluation space has data. In our experiments, combinations of complementary data augmentation algorithms were also examined. This enabled us to see the big picture of the evaluation space. 

To arrive at the results in Tables \ref{tab:data_aug_eval} and \ref{tab:speech_data_aug_eval}, all models were trained from scratch using random seeds for at least 3 times with and without \textit{AgMax}. The best test scores in each run were averaged for the evaluation reporting. We used the default parameter initialization in PyTorch \cite{paszke2019pytorch} but we observed improvement in performance if a higher entropy Gaussian distribution is used in \textit{AgMax}. For the MI loss function, $\lambda=-1$ in our experiments.

        







From Tables \ref{tab:data_aug_eval} and \ref{tab:speech_data_aug_eval} and Figure  \ref{fig:imagenet_resnet50_90_270_top1}, we make the following general observations: 

1. \textit{AgMax} consistently improves the performance of all data augmentation algorithms either as a standalone or in combination with other methods. For ResNet50 (90 epochs ImageNet in Figure \ref{fig:imagenet_resnet50_90_270_top1}) and LeNet5 (30 epochs Speech Commands in Table \ref{tab:speech_data_aug_eval}), only \textit{AgMax} achieves significant generalization improvements.

2. There is no data augmentation algorithm, separately or in combination with AutoAugment and/or \textit{AgMax} that can outperform all other methods in all datasets, models, and training conditions.  This means that there is no single superior data augmentation method among the techniques that we evaluated.

3. AutoAugment improves the performance of baseline and regional dropout algorithms. Exceptions are on ResNet50 at 90 epochs and on CutMix at 270 epochs. Both configurations were not explored in their original papers.

4. Similar to policy-based methods, \textit{AgMax} can be applied as an add-on regularizer to improve model generalization. This simplifies the overall optimization process. 

\subsection{CIFAR10 and CIFAR100}
Both CIFAR10 and CIFAR100 datasets \cite{krizhevsky2009learning} have 60,000 real-world color images of size $32\times{32}$ pixels. Both datasets have train-test split of 50,000-10,000. CIFAR10 has 10 classes while CIFAR100 is made of 100 classes.

All regularization methods were evaluated after training WideResNet28-10 and WideResNet40-2 models \cite{zagoruyko2016wide} with an initial learning rate of 0.1 for 200 epochs using cosine learning rate decay, batch size of 128, and SGD optimizer with weight decay of $5e^{-4}$. The standard data augmentation is made of random cropping of $32\times{32}$ pixels with padding size of 4 pixels, random horizontal flipping, and normalization. CutOut size is $16\times{16}$ pixels. MixUp has $\alpha{=}1$. CutMix probability is 0.5 with Beta distribution of $\alpha{=}1$.

Table \ref{tab:data_aug_eval} shows that 2 out of 4 model and dataset configurations, \textit{CutMix+AA+AgMax} is the top performing. The rest is split between \textit{CutOut+AA+AgMax} and \textit{MixUp+AA+AgMax}.

\begin{table*}
\small
\begin{center}
\begin{small}
\setlength{\tabcolsep}{3pt}

\begin{tabular}{l | r | r r r | r r r r | r r r r | r r r r }

\hline
Data & {} & \multicolumn{3}{c|}{Noise$\downarrow$} &  \multicolumn{4}{c|}{Blur$\downarrow$} &  \multicolumn{4}{c|}{Weather$\downarrow$} &  \multicolumn{4}{c}{Digital$\downarrow$}  \\
Augmentation & mCE$\downarrow$ & \footnotesize{Gauss} & \footnotesize{Shot} & \footnotesize{Impulse} & \footnotesize{Defocus} & \footnotesize{Glass} & \footnotesize{Motion} & \footnotesize{Zoom} & \footnotesize{Snow} & \footnotesize{Frost} & \footnotesize{Fog} & \footnotesize{Bright} & \footnotesize{Contrast} & \footnotesize{Elastic} & \footnotesize{Pixel} & \footnotesize{JPEG}  \\

\hline
\multicolumn{17}{c }{\textbf{without AgMax}}\\
\hline
Standard & 74.8  & 71.8 & 73.3 & 76.5 & 79.2 & 91.0 & 82.3 & 80.9 & 74.3 & 73.0 & 61.6 & 57.8 & 65.2 & 89.1 & 71.7 & 73.9     \\
CutOut & 75.2 & 74.3 & 76.4 & 80.8 & 77.3 & 91.2 & 79.1 & 79.8 & 75.5 & 74.0 & 63.3 & 57.1 & 64.7 & 87.9 & 73.4 & 73.8     \\
MixUp & 69.9 & 65.0 & 68.7 & 70.1 & 76.8 & 90.2 & 78.7 & \underline{77.6} & 67.6 &  \underline{62.7} & 54.9 & 53.9 & 54.5 & \underline{\textbf{87.6}} & 68.7 & 71.9      \\
CutMix & 75.0 & 74.5 & 76.8 & 81.8 & 79.0 & 92.3 & 81.2 & 78.2 & 73.5 & 73.8 & 62.0 & 55.9 & 64.8 & 88.9 & 68.9 & 73.9        \\
AA  & 72.6 & 66.6 & 67.1 & 71.5 & 77.6 & 88.8 & 78.1 & 83.2 & 72.9 & 73.2 & 60.3 & 54.2 & 60.2 & 91.0 & 71.9 & 72.5        \\
CutOut+AA  & 72.2 & 67.6 & 68.6 & 72.0 & 75.4 & 90.0 & 78.6 & 81.8 & 73.2 & 74.0 & 59.3 & 54.2 & 58.6 & 92.6 & \underline{68.3} & 68.5        \\

MixUp+AA & \underline{67.6} & \underline{59.8} & \underline{60.6} & \underline{63.8} & \textbf{74.9} & \textbf{86.9} & \textbf{74.9}  & 78.2 & \underline{66.4} & 64.7 & \underline{53.3} &  \underline{52.3} & \underline{53.3} & 88.6 & 69.0 & \textbf{67.1}       \\
CutMix+AA & 72.2 & 66.4 & 67.0 & 72.5 & 76.7 & 91.5 & 77.5 & 78.1 & 73.3 & 72.8 & 58.8 & 53.3 & 60.1 & 91.1 & 71.7 & 72.3         \\
\hline
\multicolumn{17}{c }{\textbf{with AgMax}}\\
\hline
Standard & 74.9 & 72.4 & 74.6 & 77.9 & 78.5 & 89.9 & 81.3 & 82.4 & 75.0 & 73.3 & 62.3 & 56.3 & 64.3 & 89.7 & 70.8 & 74.3     \\
CutOut & 74.9 & 72.7 & 75.1 & 78.6 & 78.4 & 91.8 & 79.5 & 79.9 & 74.5 & 74.5 & 61.7 & 55.8 & 63.6 & 89.5 & 74.6 & 73.6     \\
MixUp & 69.9 &  65.0 & 68.7 & 69.0 & 76.8 & 90.4 & 77.8 & \textbf{77.3} & 67.8 & 61.2 & 54.9 &  54.4 & 56.3 & 89.0 & \textbf{67.9} & 72.1         \\
CutMix  & 75.3 & 74.5 & 76.7 & 80.4 & 78.2 & 91.7 & 79.0 & \textbf{77.3} & 74.4 & 74.2 & 62.0 & 56.2 & 64.8 & 89.6 & 74.9 & 76.1         \\
AA  & 73.7 & 68.7 & 69.6 & 73.9 & 79.3 & 88.7 & 79.4 & 83.5 & 73.8 & 73.3 & 62.2 &  55.5 & 59.4 & 92.5 & 72.0 & 74.3   \\
CutOut+AA  & 73.3 & 68.4 & 69.0 & 74.6 & 77.1 & 90.6 & 78.2 & 83.6 & 75.0 & 74.5 & 61.3 &  54.1 & 58.8 & 92.5 & 71.7 & 70.8      \\
MixUp+AA  & \textbf{67.1} & \textbf{57.2} & \textbf{58.5} & \textbf{60.8} & 75.8 & 89.6 & 76.2 & 78.7 & \textbf{64.4} & \textbf{60.2} & \textbf{52.1} &  \textbf{51.9} & \textbf{52.5} & 90.4 & 69.1 & 69.8       \\
CutMix+AA & 72.9 &  70.8 & 71.5 & 75.4 & 75.8 & 91.9 & 77.8 & 78.6 & 72.2 & 72.9 & 58.1 & 52.8 & 58.9 & 90.0 & 74.0 & 72.9        \\

\hline
\end{tabular}
\vskip 0.1in
\caption{Evaluation landscape showing corruption robustness of ResNet50 model trained for 270 epochs using different regularizer configurations. mCE is Top-1\% mean Corruption Error. }
\label{tab:corruption_robustness}
\end{small}
\end{center}
\vskip -0.1in
\end{table*}

\begin{table}
\small
\begin{center}
\setlength{\tabcolsep}{0.5em}
\begin{tabular}{l | c | c  r  r }
\hline
{} & {} & {Stand-} & {} & Clean \\
{Method} & {Year} & {alone?} & {mCE$\downarrow$} & {Error$\downarrow$} \\
\hline

DeepAugment+AugMix\cite{hendrycks2020many} & 2020 & No & 53.6 & 24.2 \\
Assemble-ResNet50\cite{lee2020compounding} & 2020 & No & 56.5 & 17.9 \\
ANT $(3\times{3})$\cite{rusaksimple} & 2020 & Yes & 63.0 & 23.9 \\
BlurAfterConv \cite{vasconcelos2020effective} & 2020 & Yes	& 64.9 & 21.2 \\
AugMix\cite{hendrycks2019augmix} & 2020 & Yes & 65.3 & 22.5 \\
\hline
\textit{MixUp+AA+AgMax} & - & No & 67.1 & 21.4 \\
\hline
Stylized ImageNet\cite{geirhosimagenet} & 2019 & Yes & 69.3 & 25.1 \\
Patch Uniform\cite{lopes2019improving} & 2019 & Yes & 74.3 & 24.5 \\
Baseline & - & N/A & 76.7 & 23.8 \\

\hline
\end{tabular}
\vskip 0.1in
\caption{ImageNet-C robustness leaderboard with a ResNet50 backbone. Standalone indicates whether the method is a combination of techniques or a single method. Clean Error is the classification error on the uncorrupted validation set.}
\label{tab:mce_leaderboard}
\end{center}
\vskip -0.1in
\end{table}

\begin{table}
\small
\begin{center}

\setlength{\tabcolsep}{0.5em}
\begin{tabular}{l | r  r  r   }
\hline
Data & \multicolumn{3}{c}{FGSM$\uparrow$}  \\
Augmentation &  $\epsilon=0.1$  & $\epsilon=0.3$ & $\epsilon=0.5$   \\
\hline
\multicolumn{4}{c }{\textbf{without AgMax}}\\
\hline

Standard & 24.9 & 13.4 & 8.0 \\
CutOut & 24.6 &  12.8 & 7.4 \\
MixUp &  31.8 &  21.0 & 15.1 \\
CutMix & 34.5 & 20.4 & 13.0 \\
AA&  29.3 & 20.4 & 15.3 \\
CutOut+AA & 28.9 & 20.2 & 14.4 \\
MixUp+AA & 35.0 & 26.4 & \underline{20.7} \\
CutMix+AA & \underline{37.0} & \underline{27.2} & 20.3 \\
\hline
\multicolumn{4}{c }{\textbf{with AgMax}}\\
\hline
Standard & 28.5 & 16.6 & 10.5 \\
CutOut & 26.2 & 14.3 & 9.0 \\
MixUp & 35.0 & 23.4 & 16.4 \\
CutMix & 37.5 & 22.3 & 13.5 \\
AA & 28.1 & 19.9 & 14.9 \\
CutOut+AA & 32.0 & 22.1 & 14.4 \\
MixUp+AA & 35.6 & \underline{\textbf{28.1}} & \underline{\textbf{23.5}} \\
CutMix+AA & \underline{\textbf{37.7}} & 28.0 &  21.4 \\

\hline
\end{tabular}
\vskip 0.1in
\caption{Pre-trained (270 epochs) ResNet50 model Top-1\% accuracy after FGSM attack with increasing strength, $\epsilon$. }
\label{tab:fgsm_attack}
\end{center}
\vskip -0.1in
\end{table}

\begin{table*}
\small
\begin{center}
\begin{tabular}{ c c c c | c c c c | c c c c| c c c c  } 
\hline
\hline
\multicolumn{4}{c|}{Standard} & \multicolumn{4}{c|}{CutOut} & \multicolumn{4}{c|}{MixUp} & \multicolumn{4}{c}{CutMix} \\
MI & MSE & KL & CE & MI & MSE & KL & CE & MI & MSE & KL & CE & MI & MSE & KL & CE \\
\hline
77.4 & \textbf{77.5} & 77.3 & 77.4 & 79.2 & \textbf{80.0} & 78.9 & 79.2 & 79.5 & 78.7 & \textbf{79.7} & 79.3 & 80.0 & \textbf{80.1} & 80.0 & 80.0 \\
\hline
\hline
\multicolumn{4}{c|}{AA} & \multicolumn{4}{c|}{CutOut+AA} &
\multicolumn{4}{c|}{MixUp+AA} &
\multicolumn{4}{c}{CutMix+AA}\\

MI & MSE & KL & CE & MI & MSE & KL & CE &
MI & MSE & KL & CE & MI & MSE & KL & CE \\
\hline
79.2 & \textbf{79.6} & 79.0 & 79.0 & 81.1 & \textbf{81.2} & 81.1 & 80.9 & \textbf{80.7} & 80.1 & 80.5 & 80.5 & \textbf{81.3} & 80.8 & 81.2 & 80.9 \\
\hline
\end{tabular}
\vskip 0.1in
\caption{Top-1\% accuracy on CIFAR100 of WideResNet40-2 trained for 200 epochs using different agreement functions: MI, MSE, KL-divergence and cross-entropy (CE).}
\label{tab:other_agmax_cifar100}
\end{center}
\vskip -0.1in
\end{table*}

\begin{table}
\small
\setlength{\tabcolsep}{1.5pt}
\begin{center}
\begin{tabular}{c c c  | c c c | c c c | c c c} 
\hline
 \multicolumn{3}{c|}{MixUp} & \multicolumn{3}{c|}{CutMix} & \multicolumn{3}{c|}{MixUp+AA} & \multicolumn{3}{c}{CutMix+AA} \\
MI & MSE & KL & MI & MSE & KL & MI & MSE & KL & MI & MSE & KL \\
\hline
77.5 & 77.4 & \textbf{77.7} & 77.4 & 77.5 & \textbf{77.6} & \textbf{77.1} & 76.9 & 76.1 & \textbf{77.0} & 76.6 & 74.5  \\
\hline
\end{tabular}
\vskip 0.1in
\caption{Top-1\% accuracy on ImageNet of ResNet50 trained for 90 epochs using different agreement functions: MI, MSE and KL-divergence.}
\label{tab:other_agmax_imagenet}
\end{center}
\vskip -0.1in
\end{table}

\begin{table}
\small
\begin{center}
\begin{tabular}{l | c c c c c  } 
\hline
Model & Stan & AM & CM & CM+AA & CM+AA+AM  \\
\hline
RN101 & 78.1 & 79.2 & 79.8 & 80.7  & \textbf{81.2}  \\
\hline
RX200 & 68.8 & \textbf{70.1} & 65.6 & 63.5  & 65.5  \\
RY400 & 74.3 & \textbf{75.0} & 71.4 & 70.6  & 71.8  \\
\hline
EfNB0 & 75.3 & \textbf{75.7} & 72.3 & 71.7  & 71.9  \\
\hline
\end{tabular}
\vskip 0.1in
\caption{Top-1\% accuracy on ImageNet using ResNet101 (RN101) for 270 epochs, RegNetX200MF (RX200), RegNetY400MF (RY400) and EfficientNet-B0 (EfNB0) all for 100 epochs as done in \cite{radosavovic2020designing}. \textit{Legend:} Standard (Stan), \textit{AgMax-MI} (AM), AutoAugment (AA) and CutMix (CM). EfNB0-AM is \textit{AgMax}-MSE.}  
\label{tab:big_agmax_imagenet}
\end{center}
\vskip -0.1in
\end{table}


\subsection{ImageNet}
The ILSVRC ImageNet dataset \cite{russakovsky2015imagenet} has 1.2M real-world color images made of 1,000 classes for training. The validation set has 50,000 color images. The standard image processing includes resizing,  cropping to $224\times{224}$ pixels, random horizontal flipping, color jitter, lighting, and normalization.

All data augmentation methods were evaluated after training a ResNet50 network for 90 and 270 epochs. The initial learning rate is 0.1. The optimizer is SGD with a weight decay of $1e^{-4}$. The learning rate is reduced by a factor of 0.1 at 30, 60, and 80 epochs during the 90 epochs training and 75, 150, and 225 epochs during the 270 epochs training. Batch size is 256. CutOut size is $112\times{112}$ pixels as used in CutMix paper. MixUp has $\alpha{=}0.2$. CutMix probability is 1.0 with Beta distribution of $\alpha{=}1$.

Table \ref{tab:data_aug_eval} and Figure \ref{fig:imagenet_resnet50_90_270_top1} show that when ResNet50 is trained for 90 epochs, \textit{MixUp+AgMax} is the best performing method while \textit{CutMix+AA+AgMax} is the top performing algorithm for 270 epochs. When only one regularizer is combined with the standard data augmentation, only \textit{AgMax} has a significant positive gain in accuracy and therefore has the highest performance in 90 epochs.

\subsection{Speech Commands Dataset}
Google Speech Commands Dataset \cite{warden2018speech} contains 64,727 30-class utterances from 1,881 speakers. Each single-word utterance such as \textit{yes, no, up,} or \textit{down} is about one-second long. The standard audio signal preprocessing include random amplitude, pitch, and speed adjustment, stretching, time shifting, and addition of background noise. The audio signals are converted into $32\times{32}$ mel spectrogram input data.    

Table \ref{tab:speech_data_aug_eval} data augmentation methods were evaluated after training LeNet5 \cite{lecun1998gradient} and VGG11 \cite{simonyan2014very} for 30 epochs. The initial learning rate is 0.003 and adjusted by a factor of 0.1 after every 10 epochs for LeNet5 and 15 epochs for VGG11. The optimizer is Adam with a weight decay of $5e^{-4}$ for LeNet5 and $1e^{-4}$ for VGG11. The settings in CIFAR10/100 for CutOut, MixUp, and CutMix are used.

Table \ref{tab:speech_data_aug_eval} shows that \textit{CutOut+AgMax} is the best performing algorithm for LeNet5 and a tie between \textit{MixUp+AgMax} and \textit{CutMix+AgMax} for VGG11. Except for \textit{AgMax}, we observed that when used as a standalone regularizer all data augmentation algorithms fail to improve the generalization of LeNet5.

\subsection{Robustness}
Robustness has been increasing in importance as we deploy deep learning models on safety-critical applications. \textbf{Although the regularization methods that we evaluated are not consciously optimized for robustness}, it is worth knowing how our different configurations perform under data corruption and adversarial attack. We evaluated our ResNet50 models trained for 270 epochs using a comprehensive data corruption suite called ImageNet-C \cite{hendrycks2018benchmarking}. For the adversarial white-box attack, we subjected our trained models under Fast Gradient Sign Method (FGSM) \cite{goodfellow2014explaining}.

The mean Corruption Error (mCE) in Table \ref{tab:corruption_robustness} is normalized with respect to AlexNet \cite{krizhevsky2017imagenet} performance on ImageNet as proposed by Hendrycks \etal \cite{hendrycks2018benchmarking}. In Table \ref{tab:corruption_robustness}, the configurations that are most robust against data corruption have one method in common - MixUp. This confirms the MixUp robustness study by Zhang \etal \cite{zhang2020does}. In many cases, AutoAugment and \textit{AgMax} further improve the corruption robustness of MixUp. The results show that while configurations with CutMix have low generalization errors, they perform poorly in terms of corruption robustness. To get an idea on how the best performing method \textit{MixUp+AA+AgMax} fares in comparison with algorithms that are optimized for robustness, we borrowed the online leaderboard of Hendrycks \etal \cite{hendrycks2018benchmarking} as shown in Table \ref{tab:mce_leaderboard}. The performance of \textit{MixUp+AA+AgMax} is competitive with the SOTA.

Table \ref{tab:fgsm_attack} shows that increasing the adversary strength $\epsilon$ on the validation set, \textit{AgMax} generally improves the robustness of all pre-trained models by a wide margin. Similar to corruption robustness results, configurations that have good performance have MixUp in common. CutMix is also exhibiting robustness against FGSM.  On the average, \textit{MixUp+AA+AgMax} is also the best performing against adversarial attack.   

\subsection{Slower Training}
Normalized to Standard model training time, \textit{AgMax} incurs a performance penalty of about $1.7\times$ mainly due to the computation of the agreement between 2 positive samples. Relative to Standard model, CutMix, MixUp, CutOut and AutoAugment have minimal impact on the training time.


\subsection{Other Agreement Functions}
Other agreement functions can be used in place of MI. Table \ref{tab:other_agmax_cifar100} shows the comparison among MI, MSE, KL-divergence and CE on CIFAR100. In self-supervised learning, BYOL uses MSE while CE is the loss function in DINO. KL-divergence is also utilized since it is a good distance measure between two probability distributions. Table \ref{tab:other_agmax_cifar100} shows that MSE is the best performing agreement function on Standard, CutOut, CutMix and AA. KL excels on MixUp. MI is the top performing method on MixUp+AA and CutMix+AA and has the highest overall performance. Table \ref{tab:other_agmax_imagenet} shows that an ablation study on ResNet50, comparable results among MI, MSE and KL on MixUp and CutMix are observed. However, KL has a significantly lower performance on both MixUp+AA and CutMix+AA.  

\subsection{Other SOTA Models}
Table \ref{tab:big_agmax_imagenet} shows that other SOTA models benefit from \textit{AgMax}. Large model such as ResNet101 gains +3.1\% with CutMix+AA+\textit{AgMax}. For RegNet \cite{radosavovic2020designing} and EfficientNet \cite{tan2019efficientnet}, the use of \textit{AgMax} improves the model accuracy. However, unlike in previous models, CutMix and AA perform poorly.

\subsection{Object Detection and Instance Segmentation}
Using the MMDetection framework \cite{mmdetection}, the performance of our pre-trained ResNet50 models on object detection and instance segmentation tasks on both PascalVOC \cite{Everingham10,Everingham15} and MS COCO datasets \cite{lin2014microsoft} can be evaluated. Table \ref{tab:detection_segmentation} shows that the backbone models pre-trained with \textit{AgMax} consistently outperforms models without it. Similar to the results in robustness, models pre-trained with MixUp exhibit the best results in both detection and segmentation. Training Faster R-CNN \cite{ren2015faster} and Mask R-CNN \cite{he2017mask} models uses the default MMDetection configurations with gradient clipping and $1\times$ schedule.

\begin{table}
\small
\setlength{\tabcolsep}{3pt}
\begin{center}
\begin{tabular}{l | r | r r | r r | r r} 
\hline
{} & Stan- & \multicolumn{2}{c|}{CO+AA} & \multicolumn{2}{c|}{MU+AA}& \multicolumn{2}{c}{CM+AA}  \\
Dataset & dard & {} & \multicolumn{1}{c|}{\scriptsize{\textit{+AgMax}}} & {} & \multicolumn{1}{c|}{\scriptsize{\textit{+AgMax}}} & {} & \multicolumn{1}{c}{\scriptsize{\textit{+AgMax}}}\\
{} & \footnotesize{76.8} & \footnotesize{77.9} & \footnotesize{78.2} & \footnotesize{78.3}  & \footnotesize{78.6} & \footnotesize{78.5} & \footnotesize{79.1} \\
\hline
\multicolumn{1}{c}{} & \multicolumn{6}{c}{Faster R-CNN \cite{ren2015faster}}\\
\hline
PascalVOC & 80.5 & 81.2 & 81.4 & 81.1 & \textbf{81.5}  & 80.8 & 81.4 \\
\hline
\multicolumn{1}{c}{} & \multicolumn{6}{c}{Mask R-CNN \cite{he2017mask}}\\
\hline
COCO \textit{box} & 34.6 & 35.6 & 36.6 & 36.0 & \textbf{37.0} & 36.0 & 36.4 \\
\hline
COCO \textit{mask} & 31.3 & 32.1 & 32.8 & 32.4 & \textbf{32.9} & 32.3 & 32.6 \\
\hline
\end{tabular}
\vskip 0.1in
\caption{mAP on object detection and segmentation tasks using  ResNet50-FPN backbone trained for 270 epochs with Top-1\% accuracy indicated for reference. CO is CutOut. MU is MixUp. CM is CutMix. AA is AutoAugment.}
\label{tab:detection_segmentation}
\end{center}
\vskip -0.1in
\end{table}

\section{Conclusion}
\textit{AgMax} is a simple regularization technique that maximizes the agreement between the predictions of two positive samples. Empirical results demonstrated significant gains in performance on classification, object detection and segmentation. 

\section{Acknowledgement}
This work was funded by the UP ECWRG 2019-2020, CHED-PCARI and ERDT-FRDG. Thanks to CNL and PCARI-PRIME: Roel Ocampo and Vladimir Zurbano, for server hosting. 

{\small
\bibliographystyle{ieee_fullname}
\bibliography{egbib}
}

\end{document}